%% file: main.tex
\documentclass[sigconf]{acmart}
\usepackage{soul}
\usepackage{multirow}
\usepackage{enumitem}
\usepackage{makecell}
\usepackage{graphicx}
\usepackage{cancel}
\usepackage[normalem]{ulem}
\usepackage{colortbl}
\usepackage{xcolor}

\definecolor{grey}{rgb}{0.8,0.8,0.8}

\acmConference[Conference acronym 'XX]{Make sure to enter the correct
  conference title from your rights confirmation email}{June 03--05,
  2018}{Woodstock, NY}

\acmBooktitle{Woodstock '18: ACM Symposium on Neural Gaze Detection,
 June 03--05, 2018, Woodstock, NY} 
\acmISBN{978-1-4503-XXXX-X/18/06}

\begin{document}

\title{CoRCi: Cross-Reconstruction of Coherent Interests Modeling in Cross-Domain Sequential Recommendation}

\author{Qingtian Bian}
\affiliation{%
  \institution{Nanyang Technological University}
  \country{Singapore}}
\email{bian0027@e.ntu.edu.sg}

\author{Tieying Li}
\affiliation{%
  \institution{Northeastern University}
  \country{China}}
\email{tieying@stumail.neu.edu.cn}

\author{Marcus de Carvalho}
\affiliation{%
  \institution{Nanyang Technological University}
  \country{Singapore}}
\email{ivsucram@gmail.com}

\author{Jiaxing Xu}
\affiliation{%
  \institution{Nanyang Technological University}
  \country{Singapore}}
\email{jiaxing003@e.ntu.edu.sg}

\author{Hui Fang}
\affiliation{%
  \institution{Shanghai University of Finance and Economics}
  \country{China}}
\email{fang.hui@mail.shufe.edu.cn}

\author{Yiping Ke}
\affiliation{%
  \institution{Nanyang Technological University}
  \country{Singapore}}
\email{ypke@ntu.edu.sg}

\begin{abstract}
Cross-Domain Sequential Recommendation (CDSR) aims to alleviate data sparsity by transferring dynamic user interests across related domains. A key challenge lies in effectively bridging these domains. In single-domain modeling, models cannot distinguish between domain-specific and domain-invariant interests. Recent methods merge domain-specific sequences chronologically into a mixed-domain sequence to capture domain-invariant knowledge. However, they typically deploy separate encoders for the mixed-domain sequence and train them with per-domain loss aggregation. This workflow magnifies inter-domain discrepancies and disrupts domain-invariant interest coherence, especially when query–target pairs in Seq2Seq originate from different domains. In this paper, we present \textbf{CoRCi} (Cross-Reconstruction for Coherent Interest), a dual-target CDSR framework that tackles these drawbacks. Specifically, CoRCi proposes a Cross-Reconstruction approach that generates mixed-domain representations directly from pre-encoded specific-domain representations via cross-attention. The generated representations are then trained using a single, sequence-level, domain-agnostic loss to preserve the coherence of domain-invariant interests. To further suppress domain discrepancies in mixed-domain modeling, CoRCi introduces FocalNCE, which embeds Focal Loss into the preceding mixed-domain InfoNCE objective. The new loss assigns higher penalties to negatives drawn from the same domain as the query, thereby strengthening domain-invariant alignment. Extensive experiments on four real-world datasets demonstrate that CoRCi consistently outperforms state-of-the-art CDSR counterparts, achieving statistically significant gains across all metrics. Code is available at: \url{https://github.com/DiMarzioBian/CoRCi/}.
\end{abstract}

\ccsdesc[500]{Information systems~Recommender systems}

\keywords{Recommender System, Sequential Recommendation, Cross-Domain Sequential Recommendation, Cross Reconstruction}

\maketitle

\input{ch1_intro}

\input{ch2_related_works}

\input{ch3_preliminaries}

\input{ch4_methodology}

\input{ch5_experiment}

\input{ch6_conclusion}

\bibliographystyle{ACM-Reference-Format}
\bibliography{citation}


\end{document}

%% file: ch1_intro.tex
%
\section{Introduction} \label{sec: intro}
%
Cross-Domain Sequential Recommendation (CDSR) has emerged as a promising research area that aims to mitigate data sparsity by transferring dynamic knowledge from source domains to related target domains. The central premise of CDSR is that certain dynamic user interests are domain-invariant and evolve across multiple domains. Leveraging these shared interests not only facilitates the adaptation of existing users to new domains but also enables a more comprehensive user profiling, ultimately enhancing recommendation performance across all domains. To bridge isolated domains, recent CDSR works construct mixed-domain sequences (also known as cross-domain sequences or domain-hybrid sequences) by reorganizing interactions from each specific domain in a chronological order \cite{cao2022contrastive, ye2023dream, lin2024mixed, bian2025abxi, park2024pacer, park2023cracking}. 

\begin{figure}[t]
\centering
\includegraphics[width=1.0\linewidth]{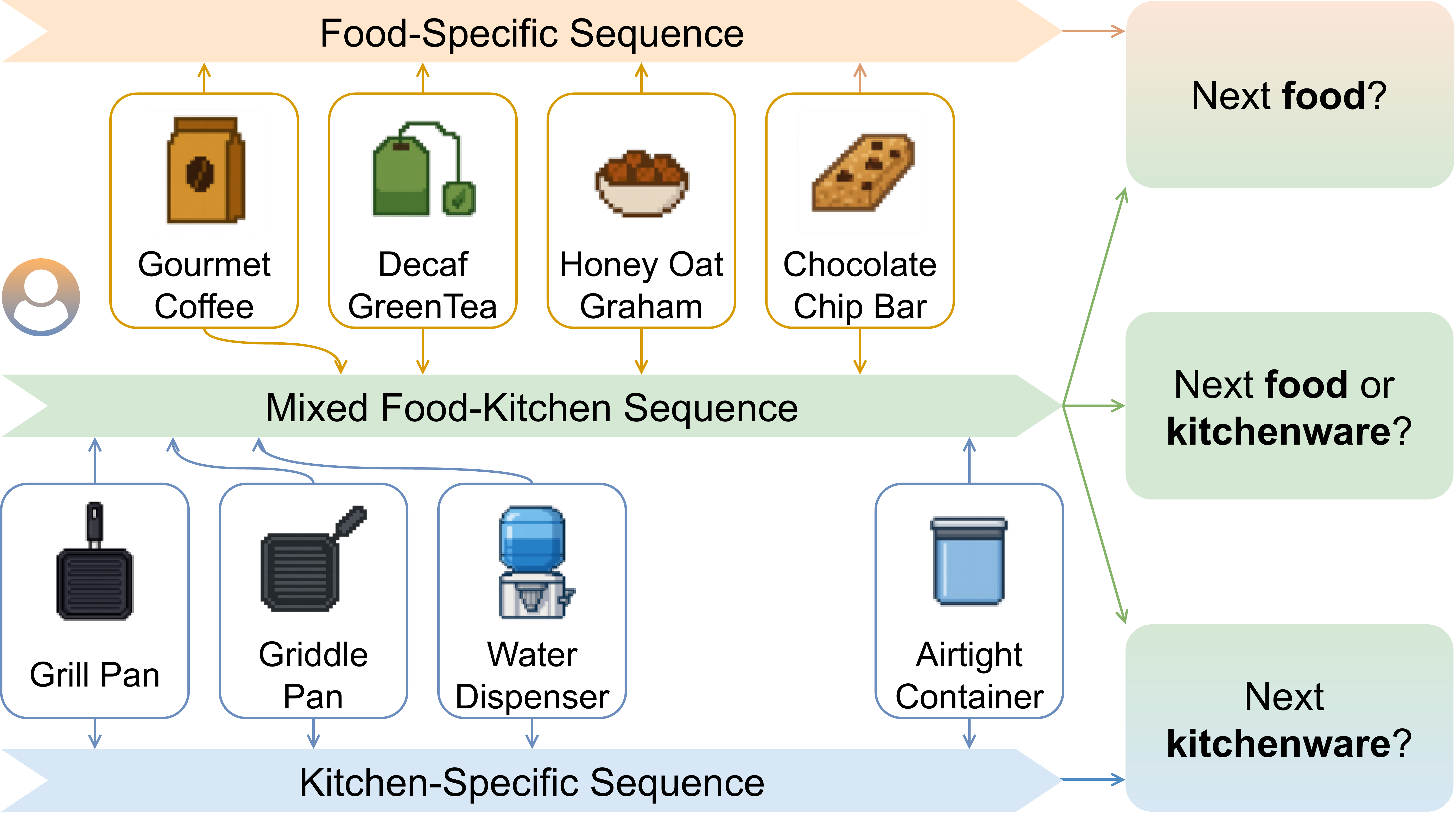}
\setlength{\abovecaptionskip}{-0.3cm}
\caption{Dual-target CDSR example on Food-Kitchen domains by user \textit{A11FFLD0GV82CQ} from the Amazon dataset.}
\label{fig: cdsr}
\vspace{-4mm}
\end{figure}
Figure~\ref{fig: cdsr} illustrates a snippet of a real user's interaction sequences (reviewerID: \textit{A11FFLD0GV82CQ}) spanning the food domain (\textit{Grocery and Gourmet Food}) and the kitchen domain (\textit{Home and Kitchen}) from the Amazon review dataset~\cite{mcauley2015image, he2016ups}. In the mixed-domain sequence, the user first purchased a \textit{Grill Pan} and a \textit{Griddle Pan}, reflecting an interest in cooking. Subsequently, the user bought a \textit{Water Dispenser}, \textit{Gourmet Coffee}, and \textit{Decaffeinated Green Tea}, indicating an interest in healthy beverages. This was followed by purchases of \textit{Honey Oat Graham}, \textit{Chocolate Chip Bars}, and an \textit{Airtight Container}, suggesting an interest in snacks and related storage solutions. These transitions demonstrate that the user's interests, particularly in beverages and then in snacks, evolve coherently between both domains. Without the mixed-domain perspective, its kitchen-specific sequence will appear fragmented and erratic. In contrast, modeling user behavior through a coherent cross-domain view enables more accurate capture of its evolving latent interests.

\begin{figure}[t]
\centering
\includegraphics[width=\linewidth]{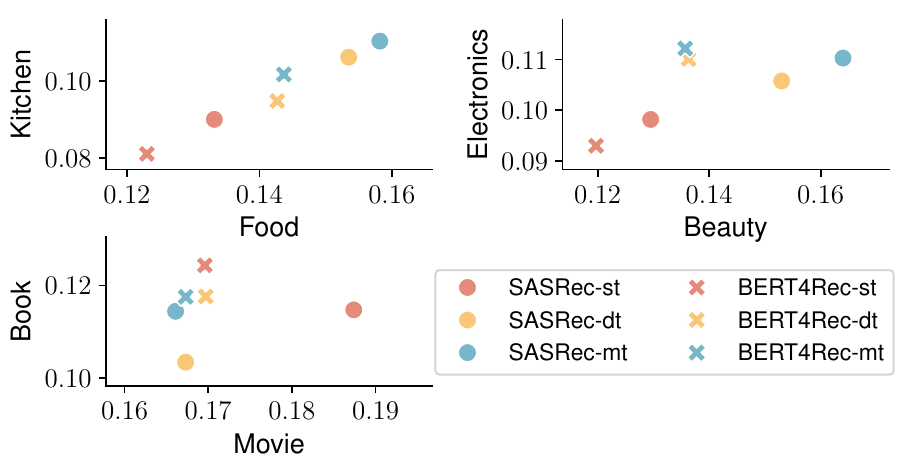}
\setlength{\abovecaptionskip}{-0.3cm}
\caption{MRR of SASRec and BERT4Rec variants in single-target (st), dual-target (dt), and mixed-target (mt) settings.}
\label{fig: preliminaries}
\vspace{-0.6cm}
\end{figure}
However, existing CDSR methods commonly employ a dual-target (DT) supervision strategy~\cite{bian2025abxi, lin2024mixed, cao2022contrastive, ye2023dream}, where token-wise losses for the mixed-domain sequence are aggregated by each specific domain (i.e., averaged per domain and summed all). This isolation undermines the inherent coherence of the mixed-domain sequence, fragmenting user interests across domains and weakening the modeling of domain-invariant interests.

To quantitatively examine the significance of preserving interest coherence through losses, we conduct a comparative analysis using two representative sequential recommendation models, SASRec~\cite{kang2018self} and BERT4Rec~\cite{sun2019bert4rec}, under three different training strategies. Single-Target (ST): models are trained independently on each specific domain. Dual-Target (DT): models are trained on mixed-domain sequences, but losses are aggregated separately for each domain. Mixed-Target (MT): models are trained on mixed-domain sequences, and the losses are aggregated without differentiating domains. Following the experimental setup in~\cite{bian2025abxi}, we evaluate these strategies on three domain pairs from the Amazon review dataset: Food–Kitchen, Beauty–Electronics, and Movie–Book. As illustrated in Figure~\ref{fig: preliminaries}, ST excels when user interests are primarily domain-specific, while MT outperforms others when domain-invariant interests are more pronounced. However, the DT models perform moderately among the three, confirming that per-domain loss aggregation is not the optimal choice.

In addition, most prior CDSR models \cite{lin2024mixed, cao2022contrastive, ye2023dream, zheng2022ddghm, xu2025heterogeneous} introduce separate encoders that learn mixed-domain representations from scratch. Within their mixed-domain Seq2Seq setup, the queries and their ground-truth targets often come from different domains, which degrades performance due to the domain gaps. ABXI~\cite{bian2025abxi} adopts LoRA to modulate a shared encoder across domains, yet it still requires additional processing to align specific-domain sequences with mixed-domain targets. Overall, current works lack a straightforward and efficient method for mitigating domain discrepancies along the mixed-domain pathway.

Therefore, we present \ul{C}r\ul{o}ss-\ul{R}econstruction for \ul{C}oherent \ul{I}nterest (\textbf{CoRCi}) to address these problems. Rather than instantiating separate encoders trained on mixed-domain sequences from scratch, CoRCi proposes a Cross-Reconstruction (CR) approach that generates mixed-domain representations directly from pre-encoded specific-domain representations using cross-attention. This approach enables CoRCi to reconnect coherent domain-invariant interests that are otherwise dispersed across specific-domain sequences. To prevent this coherence in the mixed-domain representations from being broken, CoRCi introduces FocalNCE, which integrates Focal Loss into the mixed-domain InfoNCE objective in a domain-agnostic manner. Specifically, since the mixed-domain negative samples are drawn from each specific domain, FocalNCE assigns higher penalties to those originating from the same domain as the positive query. These designs jointly mitigate the intra-domain bias that would otherwise distort the mixed-domain representation space by overemphasizing domain-specific patterns. As a result, CoRCi is better guided to capture coherent domain-invariant user interests.

We conduct extensive experiments on four real-world datasets to evaluate CoRCi against state-of-the-art CDSR methods. The results show that CoRCi consistently outperforms its counterparts across all metrics, with statistical significance (p < 0.01) in both domains of all datasets. The contributions of this paper are summarized as follows:
\begin{itemize}[leftmargin=0.4cm]

\item We propose a novel Cross-Reconstruction approach that generates mixed-domain representations from pre-encoded specific-domain sequences, thereby better preserving the coherence of domain-invariant interest in the mixed-domain sequences.

\item We introduce FocalNCE, which embeds Focal Loss into the mixed-domain InfoNCE to mitigate intra-domain bias and prevent coherent domain-invariant interests from being fragmented.

\item We conduct extensive experiments on four real-world datasets. Results show that CoRCi consistently outperforms state-of-the-art CDSR counterparts on all metrics with statistical significance.

\end{itemize}

%% file: ch2_related_works.tex
\begin{figure*}[ht]
\centering
\setlength{\abovecaptionskip}{-2mm}
\includegraphics[width=\linewidth]{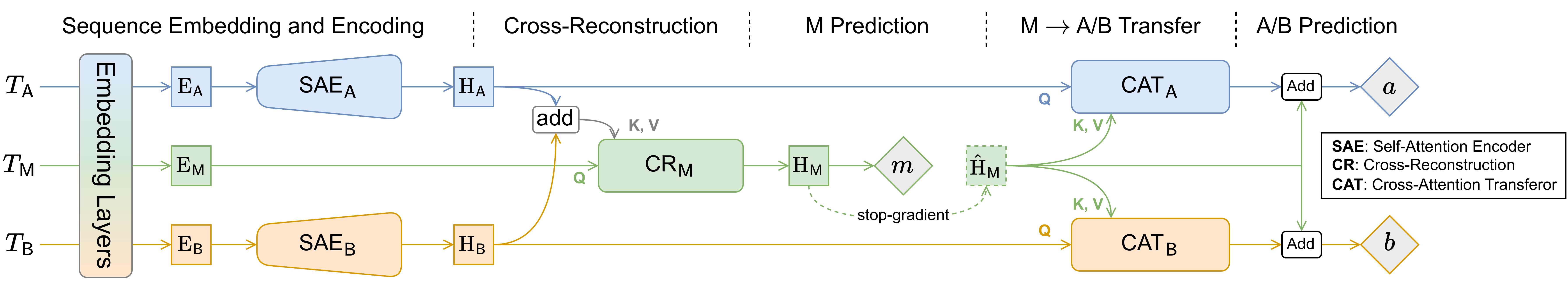}
\caption{Proposed CoRCi model.}
\label{fig: model}
\vspace{-2mm}
\end{figure*}
%

\section{Related Work}
\noindent\textbf{Sequential Recommendation (SR)} has received significant attention for modeling dynamic user preferences. Early works employed Markov Chains to model sequential interaction transitions \cite{rendle2010factorization, he2016fusing, cai2017spmc}. The emergence of deep learning brought SR with Recurrent Neural Networks (RNNs)~\cite{hidasi2015session, donkers2017sequential, wu2017recurrent}, Convolutional Neural Networks (CNNs)~\cite{zheng2017joint, tang2018personalized, yan2019cosrec}, and Graph Neural Networks (GNNs)~\cite{wu2019session, wang2020global, chang2021sequential, bian2023cpmr}. Self-attention mechanisms further advanced SR by selectively focusing on relevant interactions \cite{kang2018self, sun2019bert4rec, li2020time}. Recently, LLM-based models have demonstrated strong performance due to their superior sequence modeling capabilities \cite{harte2023leveraging, zheng2024harnessing, li2023prompt}.

\noindent\textbf{Cross-Domain Recommendation (CDR)}  exploits commonalities across domains to improve recommendation accuracy. As \cite{zhang2025comprehensive} suggests, existing CDR methods fall into three categories: mapping methods, which learn transfer functions between source and target domains \cite{finn2017model, hu2018conet, zhu2021transfer, kang2019semi}; integration methods, which combine interaction data from multiple domains to capture user preferences \cite{sheng2021one, shen2021sar, jiang2022adaptive, cao2022disencdr, zhu2023domain}; and universal recommender systems that aim to transfer domain-invariant information across domains \cite{cao2023towards, hao2024motif, zhang2024preference}.

\noindent\textbf{Cross-Domain Sequential Recommendation (CDSR)} extends CDR by incorporating temporal dynamics to transfer dynamic knowledge between domains. Some works address shared-account CDSR, where multiple anonymous users share one account \cite{ma2019pi, sun2021parallel, ma2022mixed, guo2021gcn, guo2022time}. Most studies are investigating the more general CDSR task, arguing that certain user interests are domain-invariant and can be transferred across domains to improve performance. Approaches to bridge specific domains include direct source-to-target transfer \cite{li2022recguru, alharbi2021cross, alharbi2022cross, zhuang2018cross, 10825301}, and user-mediated transfer \cite{zhang2023towards, ding2023tpuf, xu2024transfer, xu2025heterogeneous, li2023one}. Another common approach is to construct mixed-domain sequences by merging separate specific-domain sequences to better capture the domain-invariant interests. Among them, some works leverage graphs topology \cite{zheng2022ddghm, cao2022contrastive, xu2024rethinking, xu2023multi, zhang2024feddcsr}, and some works leverage self-attention \cite{li2021dual, cao2022contrastive, ye2023dream, park2024pacer, bian2025abxi, lin2024mixed, ma2024triple, park2023cracking, zheng2025fedcsr}. Recently, LLM-based approaches have also been explored to enhance CDSR~\cite{xin2025llmcdsr, liu2025bridge, shen2024exploring}.

%% file: ch3_preliminaries.tex
\section{Preliminaries}\label{sec: preliminary}
\subsection{Multi-Head Attention and Attention Block} \label{sec: attn}
Multi-head attention (MHA) \cite{vaswani2017attention} is a core technique in modern sequence models. It enables parallelized, context-aware modeling. However, the definitions and architectures associated with encoders and decoders differ slightly in various works. For instance, the Transformer's encoder does not employ the causal mask, and its decoder uses two MHA layers \cite{vaswani2017attention}. Recent LLMs adopt a decoder-only structure, where each decoder consists of one MHA layer with causal mask \cite{touvron2023llama, achiam2023gpt, bai2023qwen}, and a Feedforward network (FFN). SASRec employs MHA with the causal mask \cite{kang2018self}, and most MHA-based sequential recommenders leverage SASRec as their sequence encoder \cite{cao2022contrastive, ye2023dream, bian2025abxi}. In contrast, BERT4Rec \cite{sun2019bert4rec} employs bidirectional MHA without the causal mask. This section defines the attention module (ATTN), comprising an MHA layer and an FFN.

Formally, let the query, key, and value matrices be denoted by $\mathbf{Q}$, $\mathbf{K}$, $\mathbf{V}\in\mathbb{R}^{L\times d}$, where $L$ is the input length, and $d$ is the embedding dimension. The scaled dot-product attention is defined as:
\begin{equation}
    \text{Attention}\left(\mathbf{Q}, \mathbf{K}, \mathbf{V}\right) = \text{softmax} \left( \left(\mathbf{Q}\mathbf{K}^{\top}+\mathbf{M}^{c}\right) / \sqrt{d_k} \right)\mathbf{V},
\end{equation}
where $d_k$ is the dimension of the key vectors. The causal mask $\mathbf{M}^c\in\mathbb{R}^{L\times L}$ is defined as $\mathbf{M}^c_{ij} = \begin{cases} 0, &\text{if}\ j\leq i, \\ -\infty, &\text{if}\ j>i, \end{cases}$\ where $1\leq i, j\leq L$, which ensures that each position in a sequence can only attend to previous or current positions, and not to any future positions.

To extend single-head attention to multiple parallel heads, the MHA mechanism projects the inputs $\mathbf{Q}$, $\mathbf{K}$, and $\mathbf{V}$ into $h$ subspaces using learned parameter matrices $\mathbf{W}_i^Q\in\mathbb{R}^{d\times d_k}$, $\mathbf{W}_i^K\in\mathbb{R}^{d\times d_k}$, and $\mathbf{W}_i^V\in\mathbb{R}^{d\times d_v}$. The outputs of the attention heads are concatenated and sent into an output projection. Specifically, the MHA mechanism is given by:
\begin{gather}
    \text{MHA}\left(\mathbf{Q}, \mathbf{K}, \mathbf{V}\right) = \text{Concat} \left( \text{head}_1, \text{head}_2, ..., \text{head}_h\right)\mathbf{W}^O, \\
    \text{head}_i = \text{Attention} \left( \mathbf{Q}\mathbf{W}_i^Q, \mathbf{K}\mathbf{W}_i^K, \mathbf{V}\mathbf{W}_i^V \right),
\end{gather}
where $d_v=d_k=d/h$, and $\mathbf{W}^O\in\mathbb{R}^{d\times d}$ denotes the learnable output projection . 

To endow the model with nonlinearity and to consider interactions between different latent dimensions effectively, we adopt the FFN from Llama~\cite{touvron2023llama} and define it as:
\begin{equation}
    \text{FFN}\left(\mathbf{X}\right) = \left(\text{Swish}\left(\mathbf{X}\mathbf{W}_1 \right)\otimes\mathbf{X}\mathbf{W}_2\right)\mathbf{W}_3,
\end{equation}
where $\otimes$ denotes Hadamard product, $\mathbf{W}_1\in\mathbb{R}^{d\times \frac{8}{3}d}$, $\mathbf{W}_2\in\mathbb{R}^{d\times \frac{8}{3}d}$, and $\mathbf{W}_3\in\mathbb{R}^{\frac{8}{3}d\times d}$ are learnable matrices, and the Swish activation function is defined as $\text{Swish}(x)=\frac{x}{1+e^{-x}}$ \cite{touvron2023llama}.

We also add residual connection, dropout, and layer normalization into the pipeline to mitigate overfitting:
\begin{gather}
    \mathbf{H} = \text{MHA} \left(\mathbf{Q},\ \mathbf{K},\ \mathbf{V}\right), \\
    \mathbf{H} \leftarrow \text{LayerNorm}\left(\mathbf{Q} + \text{Dropout}\left(\mathbf{H} \right)\right), \\
    \mathbf{H} \leftarrow \text{LayerNorm}\left(\mathbf{H} + \text{Dropout}\left( \text{FFN} \left(\mathbf{H} \right) \right)\right).
\end{gather}
In implementation, all computations will be batchified for acceleration. Finally, the attention block ATTN can be abbreviated as:
\begin{equation}
    \mathbf{H} = \text{ATTN}\left(\mathbf{Q}, \mathbf{K}, \mathbf{V}\right).
\end{equation}
%

%% file: ch4_methodology.tex
\section{Methodology}
%
In this section, we introduce CoRCi, as illustrated in Figure~\ref{fig: model}. This section comprises six subsections: 1) problem formulation, 2) sequence formulation and embedding layers, 3) specific-domain self-attention encoder, 4) mixed-domain cross-reconstruction module, 5) cross-attention transferors, 6) loss design and optimization, and 7) complexity analysis.

%
\subsection{Problem Formulation} 
In this paper, we focus on the dual-target CDSR task, which is defined as follows.  Let $\mathcal{I}_\mathsf{A}$ and $\mathcal{I}_\mathsf{B}$ denote the complete item sets for non-overlapping domains A and B, respectively. For a user, let  its A-domain sequence $T_\mathsf{A} = (a_1, a_2, \ldots, a_{k}) \in \mathcal{I}_\mathsf{A}$, and B-domain sequence $T_\mathsf{B} = (b_1, b_2, \ldots, b_{q}) \in \mathcal{I}_\mathsf{B}$. The objective is to transfer knowledge between these sequences to generate top-N recommendations for the next items, $a_{k+1}$ and $b_{q+1}$, in their respective domains. The dual-target CDSR task can be formulated as:

\noindent\textbf{Input}: One user's specific-domain sequences, $T_{\mathsf{A}} = (a_1, a_2, \dots, a_k)$ and $T_{\mathsf{B}} = (b_1, b_2, \dots, b_q)$.

\noindent\textbf{Output}: A recommender system that estimates the next item that the user will interact with in each domain, $a_{k+1}$ and $b_{q+1}$.
\begin{figure}[t]
\centering
\includegraphics[width=\linewidth]{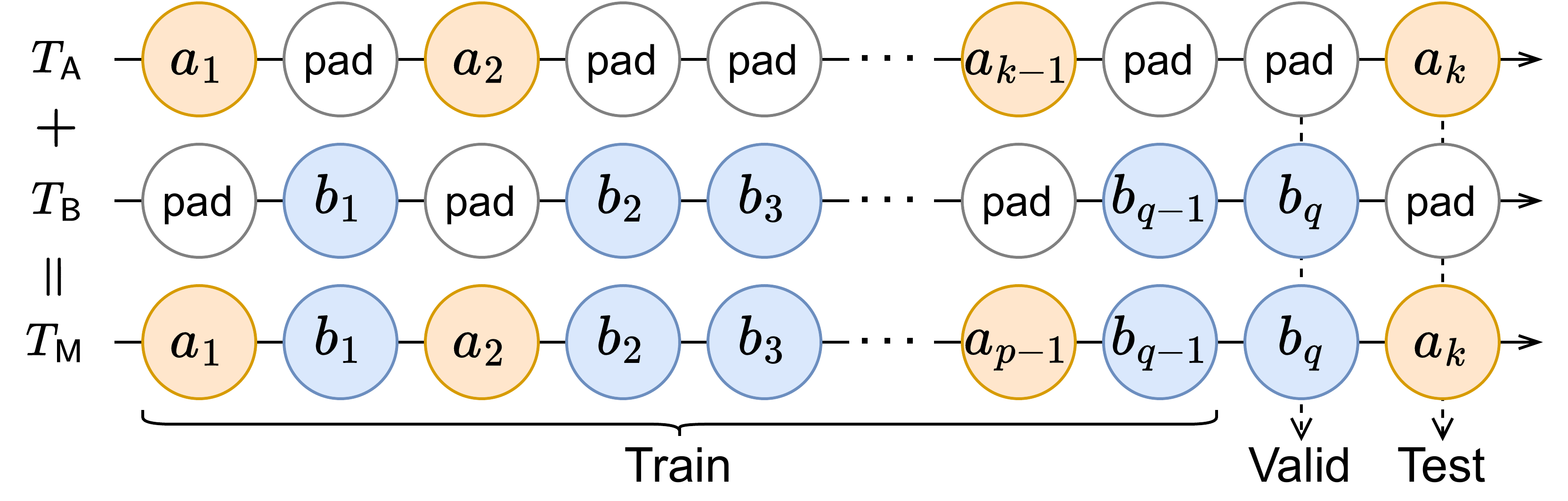}
\setlength{\abovecaptionskip}{-4mm}
\caption{An illustration of data splits of specific-domain sequences \(T_{\mathsf{A}}\) and \(T_{\mathsf{B}}\), and mixed-domain sequence \(T_{\mathsf{M}}\).}
\vspace{-4mm}
\label{fig: seq}
\end{figure}
%
\subsection{Sequence Formation and Embedding Layers}
Continuing with the above example, we define the combination of domains A and B as the mixed-domain, M. Its item set is then denoted by $\mathcal{I}_\mathsf{M} = \mathcal{I}_\mathsf{A} \cup \mathcal{I}_\mathsf{B}$. As shown in Figure~\ref{fig: seq}, we define the user's mixed-domain interaction sequence $T_\mathsf{M}$ $=$ $(a_1, b_1, a_2, b_2, b_3, \ldots, a_{k-1},\allowbreak b_{q-1})$. Every domain-specific sequences can be redefined as the domain-masked mixed-domain sequences as follows: $T_\mathsf{A}$ $=$ $(a_1, o,\allowbreak a_2, o, o, \ldots, a_{k-1}, o)$ for domain A, and $T_\mathsf{B}$ $=$ $(o, b_1, o, b_2, b_3, \ldots, o,\allowbreak b_{q-1})$ for domain B, where $o$ represents the all-zero padding item. After padding, $T_{\mathsf{A}}$ and $T_{\mathsf{B}}$ have the same length as $T_{\mathsf{M}}$ to accelerate subsequent tensor calculations. The positional index of the interactive items in each sequence will be assigned separately in reverse chronological order, ignoring padding.

We employ one shared item embedding table $\mathbf{E}^\mathcal{I}\in\mathbb{R}^{|\mathcal{I}_{\mathsf{M}}|\times d}$ and one shared positional embedding table $\mathbf{E}^\mathcal{P}\in\mathbb{R}^{L\times d}$ to project each numeric item index and positional index into a $d$-dimensional embedding vector, respectively. $L$ denotes the maximum input length. Finally, we represent the embedding of $T_{\mathsf{M}}$, $T_{\mathsf{A}}$, and $T_{\mathsf{B}}$ as $\mathbf{E}_\mathsf{M}$, $\mathbf{E}_\mathsf{A}$, and $\mathbf{E}_\mathsf{B}$, respectively, where each embedding is the summation of corresponding item embedding and positional embedding.

\subsection{Self-Attention Encoders}
Inspired by SASRec~\cite{kang2018self}, we leverage the self-attention mechanism to conduct sequential encoding. Specifically, at this stage, we instantiate one self-attention encoder (SAE) for each specific domain, A or B, respectively. The process can be described as follows:
\begin{equation} \label{eq: sae}
\begin{aligned}
    \mathbf{H}_{\mathsf{A}} &= \text{SAE}_\mathsf{A}\left( \mathbf{E}_\mathsf{A} \right) = \text{ATTN}^{\text{SAE}}_{\mathsf{A}} \left(\mathbf{E}_\mathsf{A},\ \mathbf{E}_\mathsf{A},\ \mathbf{E}_\mathsf{A} \right), \\ 
    \mathbf{H}_{\mathsf{B}} &= \text{SAE}_\mathsf{B}\left( \mathbf{E}_\mathsf{B} \right) = \text{ATTN}^{\text{SAE}}_{\mathsf{B}} \left(\mathbf{E}_\mathsf{B},\ \mathbf{E}_\mathsf{B},\ \mathbf{E}_\mathsf{B} \right).
\end{aligned}
\end{equation}
%

%
\subsection{Cross-Reconstruction}
To reconnect domain-invariant interest dispersed in specific-domain sequences, we design a cross-attention module under the Cross-Reconstruction approach, denoted CR${_\mathsf{M}}$, which produces mixed-domain representations from the specific-domain representations pre-encoded in Equation~\ref{eq: sae}. Specifically, CR$_{\mathsf{M}}$ leverages a cross-attention layer~\cite{kang2018self} to conduct the reconstruction. Because at each time step at most one of $\mathbf{H}_{\mathsf{A}}$ or $\mathbf{H}_{\mathsf{B}}$ holds a non-padding token, we use the sum of these two representations as the K and V input for cross-attention. For the Q input, we employ the $\mathbf{E}_{\mathsf{M}}$ to introduce the mixed domain position information. The overall process of CR$_{\mathsf{M}}$ can then be formulated as follows:
\begin{equation} \label{eq: cr}
\begin{aligned}
    \mathbf{H}_{\mathsf{M}} &= \text{CR}_\mathsf{M}\left(\mathbf{E}_\mathsf{M}, \mathbf{H}_\mathsf{A}, \mathbf{H}_\mathsf{B} \right) \\ 
    &= \text{ATTN}^{\text{CR}}_{\mathsf{M}} \left(\mathbf{E}_\mathsf{M},\ \mathbf{H}_\mathsf{A}+\mathbf{H}_\mathsf{B},\ \mathbf{H}_\mathsf{A}+\mathbf{H}_\mathsf{B} \right).
\end{aligned}
\end{equation}
%

%
\subsection{Cross-Attention Transferors}
To transfer the reconstructed mixed-domain knowledge back into the specific-domain representations, we again leverage cross-attention. Unlike token-wise addition, which only aligns corresponding positions, cross-attention enables each token in a specific domain to attend to the entire history of mixed-domain interactions, yielding a more comprehensive transfer. Specifically, we instantiate a Cross-Attention Transfer (CAT) for each specific domain A or B, respectively, which are formulated as follows:
\begin{equation} \label{eq: f_ab}
\begin{aligned}
    \mathbf{H}_{\mathsf{A}} &\leftarrow \text{CAT}_\mathsf{A}\left( \mathbf{H}_\mathsf{A}, \hat{\mathbf{H}}_\mathsf{M} \right) = \text{ATTN}^{\text{CAT}}_{\mathsf{A}} \left(\mathbf{H}_\mathsf{A},\ \hat{\mathbf{H}}_\mathsf{M},\ \hat{\mathbf{H}}_\mathsf{M} \right), \\
    \mathbf{H}_{\mathsf{B}} &\leftarrow \text{CAT}_\mathsf{B}\left( \mathbf{H}_\mathsf{B}, \hat{\mathbf{H}}_\mathsf{M} \right) = \text{ATTN}^{\text{CAT}}_{\mathsf{B}} \left(\mathbf{H}_\mathsf{B},\ \hat{\mathbf{H}}_\mathsf{M},\ \hat{\mathbf{H}}_\mathsf{M} \right),
\end{aligned}
\end{equation}
where $\hat{\mathbf{H}}_{\mathsf{M}}$ denotes stop-gradient $\mathbf{H}_\mathsf{M}$. This prevents mixed-domain modeling from being distorted by specific-domain supervision, thereby maintaining the coherence of domain-invariant interests.

%
\subsection{Losses and Optimization}\label{sec: loss}
%
In this subsection, we introduce the losses used for optimizing specific and mixed domains. For specific domains, we employ vanilla InfoNCE~\cite{oord2018representation}. Given a query representation $h$, its ground truth embedding $e^+$, and a candidate set $E^{\pm}$ formed by $e^+$ and the embeddings of $N_{neg}=128$ unobserved negative samples within the same domain, InfoNCE is defined as:
\begin{equation}\label{eq: infonce}
    \text{InfoNCE}\left(h,\; e^+,\; E^{\pm}\right) = - \text{log}\frac{\text{exp}\left(h \cdot e^+/\tau\right)}{\sum_{e\in E^{\pm}} \text{exp}\left(h \cdot e/\tau\right)},
\end{equation}
where $\tau$ denotes the temperature hyperparameter. Hence, the losses for domains A and B are computed as:
\begin{equation}
\begin{aligned}
    \mathcal{L}_{\mathsf{A}} &= \frac{1}{|T_{\mathsf{A}}|}\sum\nolimits_{j=1}^{|T_A|} \text{InfoNCE}\left(h_{a, j} + \hat{h}_{m, j}^{\ },\ e^+_{a, j},\ E^{\pm}_{a, j}\right), \\
    \mathcal{L}_{\mathsf{B}} &= \frac{1}{|T_{\mathsf{B}}|}\sum\nolimits_{j=1}^{|T_B|} \text{InfoNCE}\left(h_{b, j} + \hat{h}_{m, j}^{\ },\ e^+_{b, j},\ E^{\pm}_{b, j}\right),
\end{aligned}
\end{equation}
where $\hat{h}_{m, j}$ denotes the corresponding stop-gradient encoded mixed-domain representation of the same item. Here, we regard the mixed-domain embedding $\hat{h}_{m,j}$ as a \ul{task-agnostic backbone}, and the specific-domain representations derived in Equation~\ref{eq: f_ab} are then trained to learn the residuals that tailor this backbone to specific downstream objectives in each domain.

For the mixed domain, let the query representation be $h_m$. We form the candidate set as: $E^{\pm}_{\mathsf{m}} = \{e^+_{\mathsf{m}}\} \cup E^-_{\mathsf{A}} \cup E^-_{\mathsf{B}}$, where $e^+_{\mathsf{m}}$ is the embedding of the ground-truth item (positive), and $N_{\text{neg}}$ unobserved negatives are sampled uniformly from each specific domain to form the negative sets $E^-_{\mathsf{A}}$ and $E^-_{\mathsf{B}}$, respectively. Considering that items drawn from the same domain tend to have higher similarity, inner-product similarity without differentiating domains may tend to introduce intra-domain bias. To counteract this issue, we embed the Focal loss~\cite{lin2017focal} into the InfoNCE objective, proposing FocalNCE. Given $\alpha{>}0$ is the focusing parameter. FocalNCE is then defined as:
\begin{equation}
\begin{gathered}\label{eq: focal}
    p = \frac{\text{exp}\left(h \cdot e^+/\tau\right)}{\sum_{e\in E^{\pm}} \text{exp}\left(h \cdot e/\tau\right)}, \\
    \text{FocalNCE}\left(h,\; e^+,\; E^{\pm}\right) = -(1-p)^{\alpha}\text{log}(p),
\end{gathered}
\end{equation}

FocalNCE regards each similarity score as a class probability. Similarities involving the same-domain negatives constitute naturally difficult classes and are penalized extra. Consequently, FocalNCE suppresses this intra-domain bias that would otherwise skew the mixed-domain space, thereby fostering coherent domain-invariant interest representations in CoRCi. Hence, the mixed domain loss is:
\begin{equation}
    \mathcal{L}_{M} = \frac{1}{|T_M|}\sum\nolimits_{j=1}^{|T_M|} \text{FocalNCE}\left(h^F_{m, j},\ e^+_{m, j},\ E^{\pm}_{m, j}\right),
\end{equation}

Eventually, given the weight parameter $\beta$, the overall objective function is a linear combination of the losses from each domain:
\begin{equation}
    \mathcal{L} = \mathcal{L}_{\mathsf{A}} + \mathcal{L}_{\mathsf{B}} + \beta\mathcal{L}_{\mathsf{M}}. \\
\end{equation}

%
\subsection{Complexity Analysis}\label{sec: complex}
%

For FLOPs in one forward pass, the MHA module requires $4L^2d + 8Ld^2$, and the FFN requires $168Ld^2$. For ABXI, the shared SA and FFN process three sequences, and each of the three projectors (using FFN structure) processes one sequence. Therefore, the total FLOPs are 3 SA plus 6 FFN, which gives $12L^2d + 120Ld^2$. In comparison, for CoRCi, each SA and FFN processes only one sequence, resulting in 5 SA plus 5 FFN, which is $20L^2d + 120Ld^2$ around 1.06× that of ABXI. We omit normalization for both models and LoRA for ABXI. Hence, CoRCi requires a similar number of FLOPs but has more parameters, consistent with the scaling laws of LLMs and generative recommenders~\cite{zhai2024actions, zhang2024wukong}. Excluding item embeddings, the model contains a total of 5,528,624 learnable dense parameters. In terms of training time, each epoch of all datasets takes less than 30 seconds to train on an NVIDIA RTX4090.

Scaling CoRCi from $2$ to $k+2$ domains adds $k$ additional SAEs and $k$ additional CATs, while the number of CR remains fixed at one. As more domains are added, VRAM usage increases roughly proportionally, since both models follow a similar architectural paradigm.

%% file: ch5_experiment.tex
%
\section{Experiments}
In this section, we conduct extensive experiments on four publicly available real-world datasets to evaluate the effectiveness of CoRCi and aim to address the following research questions (RQ): \textbf{RQ1} How does CoRCi's performance compare to SOTA DT-CDSR models and other baselines? \textbf{RQ2} What impact do the proposed SAE-CR-CAT structure, stop-gradient operation, and the mixed-domain task have on final performance? \textbf{RQ3} How do the focusing parameter $\alpha$ of FocalNCE and the loss selections mitigates domain discrepancies? \textbf{RQ4} How do the weight $\beta$ of mixed-domain losses affect the final performance? \textbf{RQ5} How does CoRCi perform on a dataset with three domains and non-overlapping users?
\begin{table}[!t]
\caption{Statistics of CDSR Datasets.}
\vspace{-0.3cm}
\label{tab: data}
    \begin{tabular}{cccccc}
    \toprule
        Dataset & \#users & \#items & \#inter. & \#val. & \#test \\
    \midrule
        Amazon-Food & \multirow{2}*{7,144} & 11,837 & 83,663 & 2,837 & 2,419 \\
        Amazon-Kitchen & & 16,258 & 89,885 & 4,307 & 4,725 \\
    \midrule
        Amazon-Beauty & \multirow{2}*{4,474} & 10,379 & 50,329 & 2,086 & 1,875 \\
        Amazon-Elect. & & 14,188 & 63,800 & 2,388 & 2,599 \\
    \midrule
        Amazon-Movie & \multirow{2}*{28,350} & 35,712 & 347,654 & 11,728 & 10,935 \\
        Amazon-Book & & 90,958 & 403,147 & 16,622 & 17,415 \\
    \midrule
        Douban-Movie & \multirow{3}*{2,674} & 15,043 & 208,667 & 1,990 & 1,788 \\
        Douban-Book & & 6,234 & 20,942 & 471 & 639 \\
        Douban-Music & & 4,300 & 11,758 & 213 & 247 \\
    \bottomrule
    \end{tabular}
\vspace{-5mm}
\end{table}

%
\subsection{Datasets} \label{sec: dataset}
%
We conduct experiments on three two-domain datasets constructed from six subsets of the Amazon review dataset~\cite{mcauley2015image, he2016ups}. These datasets include Food-Kitchen (\textbf{AFK}: combining `Grocery and Gourmet Food' as Food and `Home and Kitchen' as Kitchen), Beauty-Electronics (\textbf{ABE}: combining `Beauty' as Beauty and `Electronics' as Electronics), and Movie-Book (\textbf{AMB}: combining `Movies and TV' as `Movie and `Books' as Book). During preprocessing, each review is treated as an interaction. First, we select users with valid interaction history in each specific domain, and merge them chronologically to construct mixed-domain sequences. Next, we filter out items with fewer than ten interactions. Following prior studies \cite{bian2025abxi, kang2018self}, we retain a maximum of 50 recent interactions for each mixed-domain sequence. Lastly, we exclude users whose specific-domain sequences contain fewer than five interactions in any domain. The statistics of these datasets are summarized in Table~\ref{tab: data}.

We train each model five times using different random seeds to assess stability and robustness. The performance is evaluated using three metrics: Hit Rate (HR), Normalized Discounted Cumulative Gain (NDCG), and Mean Reciprocal Rank (MRR).

\begin{table}[t]
\setlength{\abovecaptionskip}{0.1cm}
\caption{Hyperparameter selections.}
    \begin{tabular}{c|c}
    \toprule
    Hyperparameter & Value \\
    \midrule
    Embedding dimension $d$ & 256 \\
    ATTN layer / head $h$ & 1 / 2 \\
    Temperature $\tau$ & 0.75 \\
    Batch size $B$ & 256 (Amazon), 128 (Douban) \\
    Max / Warm-up epoch & 500 / 10 epochs \\
    Early-stopping patience & 60 stable epochs \\
    Optimizer & AdamW \\
    Learning rate & \{1e-3, 1e-4\} \\
    Weight decay & \{5, 2, 1\}$\times$\{1e1, 1e0, 1e-1, 1e-2, 1e-3\}, 0 \\
    Learning rate decay & $\times$\{0.1 to 1.0\} after 30 stable epochs \\
    Dropout rate & 0.0 to 0.9 \\
    Random seed & \{3407, 0, 1, 2, 3\} \\
    \bottomrule
    \end{tabular}
\vspace{-6mm}
\label{tab: hyper}
\end{table}
\begin{table*}[t]
\caption{Overall performance (RQ1). The best results are presented in bold, and the runner-up is underlined. Paired t-tests confirm that the pairwise differences between CoRCi and the best baseline are statistically significant (all p-values $\leq$ 0.01).}
\vspace{-0.3cm}
\small
  \begin{tabular}{cc|cccc|cccc}
    \toprule
    \multirow{2}*{Type} & \multirow{2}*{Methods} & \multicolumn{4}{c|}{Food} & \multicolumn{4}{c}{Kitchen} \\
    & & HR@5 & HR@10 & NDCG@10 & MRR & HR@5 & HR@10 & NDCG@10 & MRR \\
    \midrule
    \multirow{2}*{ST-SR} & SASRec-st & 0.1930\tiny{±0.0028} & 0.2611\tiny{±0.0036} & 0.1561\tiny{±0.0021} & 0.1332\tiny{±0.0019} & 0.1241\tiny{±0.0026} & 0.1851\tiny{±0.0018} & 0.1040\tiny{±0.0005} & 0.0900\tiny{±0.0007} \\
    & BERT4Rec-st & 0.1819\tiny{±0.0035} & 0.2528\tiny{±0.0037} & 0.1462\tiny{±0.0027} & 0.1230\tiny{±0.0030} & 0.1114\tiny{±0.0040} & 0.1685\tiny{±0.0036} & 0.0926\tiny{±0.0029} & 0.0810\tiny{±0.0025} \\
    \multirow{2}*{DT-SR} & SASRec-dt & 0.2313\tiny{±0.0034} & 0.2854\tiny{±0.0049} & 0.1797\tiny{±0.0039} & 0.1535\tiny{±0.0034} & 0.1510\tiny{±0.0037} & 0.2168\tiny{±0.0049} & 0.1248\tiny{±0.0027} & 0.1062\tiny{±0.0021} \\
    & BERT4Rec-dt & 0.2223\tiny{±0.0053} & 0.2956\tiny{±0.0030} & 0.1727\tiny{±0.0033} & 0.1427\tiny{±0.0041} & 0.1363\tiny{±0.0043} & 0.2055\tiny{±0.0052} & 0.1116\tiny{±0.0032} & 0.0948\tiny{±0.0025} \\
    \multirow{2}*{MT-SR} & SASRec-mt & 0.230\tiny{±0.0054} & 0.2905\tiny{±0.0022} & 0.1842\tiny{±0.0040} & 0.1582\tiny{±0.0048} & 0.1580\tiny{±0.0033} & 0.2251\tiny{±0.0044} & 0.1298\tiny{±0.0020} & 0.1104\tiny{±0.0015} \\
    & BERT4Rec-mt & 0.2226\tiny{±0.0052} & 0.2847\tiny{±0.0018} & 0.1716\tiny{±0.0024} & 0.1437\tiny{±0.0032} & 0.1479\tiny{±0.0033} & 0.2124\tiny{±0.0035} & 0.1200\tiny{±0.0032} & 0.1017\tiny{±0.0032} \\
    \midrule
    \multirow{3}*{ST-CDSR} & CD-SASRec & 0.1797\tiny{±0.0079} & 0.2454\tiny{±0.0046} & 0.1421\tiny{±0.0060} & 0.1197\tiny{±0.0066} & 0.1119\tiny{±0.0067} & 0.1757\tiny{±0.0070} & 0.0946\tiny{±0.0045} & 0.0821\tiny{±0.0039} \\
    & CD-ASR & 0.1976\tiny{±0.0042} & 0.2727\tiny{±0.0052} & 0.1616\tiny{±0.0028} & 0.1368\tiny{±0.0026} & 0.1345\tiny{±0.0043} & 0.1995\tiny{±0.0044} & 0.1107\tiny{±0.0037} & 0.0941\tiny{±0.0034} \\ 
    & MGCL & 0.1932\tiny{±0.0041} & 0.2673\tiny{±0.0054} & 0.1523\tiny{±0.0021} & 0.1260\tiny{±0.0018} & 0.1467\tiny{±0.0009} & 0.2157\tiny{±0.0026} & 0.1203\tiny{±0.0019} & 0.1017\tiny{±0.0019} \\
    \midrule
    \multirow{4}*{DT-CDSR} & C$^2$DSR & 0.1984\tiny{±0.0072} & 0.2574\tiny{±0.0116} & 0.1546\tiny{±0.0050} & 0.1311\tiny{±0.0035} & 0.1263\tiny{±0.0051} & 0.1879\tiny{±0.0061} & 0.1051\tiny{±0.0033} & 0.0903\tiny{±0.0027} \\
    & DREAM & 0.2158\tiny{±0.0043} & 0.2771\tiny{±0.0039} & 0.1698\tiny{±0.0025} & 0.1441\tiny{±0.0021} & 0.1377\tiny{±0.0021} & 0.2045\tiny{±0.0033} & 0.1138\tiny{±0.0012} & 0.0956\tiny{±0.0006} \\
    & ABXI & \ul{0.2498}\tiny{±0.0022} & \ul{0.3175}\tiny{±0.0041} & \ul{0.1973}\tiny{±0.0025} & \ul{0.1679}\tiny{±0.0028} & \ul{0.1737}\tiny{±0.0031} & \ul{0.2410}\tiny{±0.0026} & \ul{0.1415}\tiny{±0.0012} & \ul{0.1206}\tiny{±0.0014} \\
    & \textbf{CoRCi} & \textbf{0.2647}\tiny{±0.0011} & \textbf{0.3317}\tiny{±0.0036} & \textbf{0.2069}\tiny{±0.0023} & \textbf{0.1757}\tiny{±0.0023} & \textbf{0.1848}\tiny{±0.0033} & \textbf{0.2573}\tiny{±0.0048} & \textbf{0.1506}\tiny{±0.0023} & \textbf{0.1275}\tiny{±0.0014} \\
    \midrule
    & & \multicolumn{4}{c|}{Beauty} & \multicolumn{4}{c}{Electronics} \\
    \midrule
    \multirow{2}*{ST-SR} & SASRec-st & 0.1837\tiny{±0.0014} & 0.2597\tiny{±0.0038} & 0.1523\tiny{±0.0020} & 0.1295\tiny{±0.0016} & 0.1345\tiny{±0.0052} & 0.1894\tiny{±0.0038} & 0.1111\tiny{±0.0032} & 0.0982\tiny{±0.0033} \\
    & BERT4Rec-st & 0.1687\tiny{±0.0043} & 0.2438\tiny{±0.0043} & 0.1404\tiny{±0.0034} & 0.1197\tiny{±0.0032} & 0.1277\tiny{±0.0067} & 0.1832\tiny{±0.0069} & 0.1053\tiny{±0.0054} & 0.0930\tiny{±0.0050} \\
    \multirow{2}*{DT-SR} & SASRec-dt & 0.2292\tiny{±0.0102} & 0.3262\tiny{±0.0032} & 0.1866\tiny{±0.0032} & 0.1530\tiny{±0.0032} & 0.1481\tiny{±0.0036} & 0.2169\tiny{±0.0041} & 0.1236\tiny{±0.0040} & 0.1058\tiny{±0.0039} \\
    & BERT4Rec-dt & 0.1970\tiny{±0.0027} & 0.3077\tiny{±0.0089} & 0.1679\tiny{±0.0057} & 0.1363\tiny{±0.0050} & 0.1519\tiny{±0.0049} & 0.2246\tiny{±0.0028} & 0.1275\tiny{±0.0031} & 0.1101\tiny{±0.0031} \\
    \multirow{2}*{MT-SR} & SASRec-mt & 0.2457\tiny{±0.0095} & 0.3383\tiny{±0.0071} & 0.1983\tiny{±0.0056} & 0.1640\tiny{±0.0060} & 0.1565\tiny{±0.0032} & 0.2230\tiny{±0.0046} & 0.1281\tiny{±0.0022} & 0.1103\tiny{±0.0017} \\
    & BERT4Rec-mt & 0.2018\tiny{±0.0106} & 0.3095\tiny{±0.0115} & 0.1679\tiny{±0.0076} & 0.1357\tiny{±0.0058} & 0.1584\tiny{±0.0041} & 0.2298\tiny{±0.0011} & 0.1306\tiny{±0.0016} & 0.1122\tiny{±0.0019} \\
    \midrule
    \multirow{3}*{ST-CDSR} & CD-SASRec & 0.1605\tiny{±0.0115} & 0.2530\tiny{±0.0166} & 0.1380\tiny{±0.0107} & 0.1162\tiny{±0.0085} & 0.1290\tiny{±0.0060} & 0.1842\tiny{±0.0030} & 0.1069\tiny{±0.0027} & 0.0948\tiny{±0.0032} \\ 
    & CD-ASR & 0.1661\tiny{±0.0065} & 0.2550\tiny{±0.0044} & 0.1424\tiny{±0.0033} & 0.1211\tiny{±0.0033} & 0.1355\tiny{±0.0049} & 0.1938\tiny{±0.0039} & 0.1122\tiny{±0.0033} & 0.0987\tiny{±0.0034} \\ 
    & MGCL & 0.1364\tiny{±0.0047} & 0.2109\tiny{±0.0078} & 0.1162\tiny{±0.0044} & 0.1001\tiny{±0.0035} & 0.1537\tiny{±0.0018} & 0.2159\tiny{±0.0042} & 0.1273\tiny{±0.0019} & 0.1118\tiny{±0.0012} \\
    \midrule
    \multirow{4}*{DT-CDSR} & C$^2$DSR & 0.1835\tiny{±0.0066} & 0.2645\tiny{±0.0034} & 0.1519\tiny{±0.0038} & 0.1290\tiny{±0.0037} & 0.1288\tiny{±0.0072} & 0.1859\tiny{±0.0063} & 0.1081\tiny{±0.0047} & 0.0960\tiny{±0.0043} \\
    & DREAM & 0.2090\tiny{±0.0047} & 0.3043\tiny{±0.0069} & 0.1742\tiny{±0.0032} & 0.1447\tiny{±0.0032} & 0.1216\tiny{±0.0040} & 0.1817\tiny{±0.0041} & 0.1023\tiny{±0.0024} & 0.0895\tiny{±0.0019} \\
    & ABXI & \ul{0.2807}\tiny{±0.0082} & \ul{0.3835}\tiny{±0.0050} & \ul{0.2245}\tiny{±0.0043} & \ul{0.1846}\tiny{±0.0038} & \ul{0.1659}\tiny{±0.0021} & \ul{0.2389}\tiny{±0.0032} & \ul{0.1385}\tiny{±0.0014} & \ul{0.1200}\tiny{±0.0019} \\
    & \textbf{CoRCi} & \textbf{0.3006}\tiny{±0.0081} & \textbf{0.4020}\tiny{±0.0043} & \textbf{0.2374}\tiny{±0.0036} & \textbf{0.1946}\tiny{±0.0045} & \textbf{0.1854}\tiny{±0.0033} & \textbf{0.2630}\tiny{±0.0048} & \textbf{0.1544}\tiny{±0.0022} & \textbf{0.1331}\tiny{±0.0013} \\
    \midrule
    & & \multicolumn{4}{c|}{Movie} & \multicolumn{4}{c}{Book} \\
    \midrule
    \multirow{2}*{ST-SR} & SASRec-st & 0.2258\tiny{±0.0031} & 0.2961\tiny{±0.0037} & 0.1647\tiny{±0.0025} & 0.1874\tiny{±0.0027} & 0.1357\tiny{±0.0029} & 0.1789\tiny{±0.0033} & 0.1007\tiny{±0.0022} & 0.1147\tiny{±0.0023}\\
    & BERT4Rec-st & 0.2329\tiny{±0.0018} & 0.3105\tiny{±0.0012} & 0.1927\tiny{±0.0007} & 0.1696\tiny{±0.0007} & 0.1638\tiny{±0.0017} & 0.2152\tiny{±0.0013} & 0.1378\tiny{±0.0008} & 0.1243\tiny{±0.0006} \\
    \multirow{2}*{DT-SR} & SASRec-dt & 0.2303\tiny{±0.0046} & 0.3067\tiny{±0.0043} & 0.1903\tiny{±0.0032} & 0.1673\tiny{±0.0030} & 0.1356\tiny{±0.0015} & 0.1830\tiny{±0.0014} & 0.1146\tiny{±0.0011} & 0.1034\tiny{±0.0010} \\
    & BERT4Rec-dt & 0.2317\tiny{±0.0008} & 0.3095\tiny{±0.0011} & 0.1925\tiny{±0.0015} & 0.1697\tiny{±0.0017} & 0.1547\tiny{±0.0014} & 0.2063\tiny{±0.0012} & 0.1302\tiny{±0.0009} & 0.1176\tiny{±0.0009} \\
    \multirow{2}*{MT-SR} & SASRec-mt & 0.2284\tiny{±0.0041} & 0.3077\tiny{±0.0067} & 0.1894\tiny{±0.0044} & 0.1661\tiny{±0.0038} & 0.1516\tiny{±0.0042} & 0.2014\tiny{±0.0040} & 0.1268\tiny{±0.0031} & 0.1144\tiny{±0.0029} \\
    & BERT4Rec-mt & 0.2304\tiny{±0.0029} & 0.3074\tiny{±0.0024} & 0.1902\tiny{±0.0017} & 0.1673\tiny{±0.0016} & 0.1538\tiny{±0.0030} & 0.2072\tiny{±0.0036} & 0.1305\tiny{±0.0024} & 0.1175\tiny{±0.0019} \\
    \midrule
    \multirow{3}*{ST-CDSR} & CD-SASRec & 0.2347\tiny{±0.0022} & 0.3117\tiny{±0.0026} & 0.1940\tiny{±0.0015} & 0.1709\tiny{±0.0017} & 0.1710\tiny{±0.0042} & 0.2253\tiny{±0.0043} & 0.1434\tiny{±0.0030} & 0.1285\tiny{±0.0026} \\
    & CD-ASR & 0.232\tiny{±0.0045} & 0.3052\tiny{±0.0032} & 0.1956\tiny{±0.0027} & 0.1743\tiny{±0.0024} & 0.1622\tiny{±0.0010} & 0.2118\tiny{±0.0024} & 0.1372\tiny{±0.0010} & 0.1244\tiny{±0.0010} \\ 
    & MGCL & 0.2097\tiny{±0.0042} & 0.2851\tiny{±0.0040} & 0.1726\tiny{±0.0032} & 0.1509\tiny{±0.0033} & 0.1248\tiny{±0.0038} & 0.1668\tiny{±0.0049} & 0.1043\tiny{±0.0035} & 0.0946\tiny{±0.0033} \\
    \midrule
    \multirow{4}*{DT-CDSR} & C$^2$DSR & 0.2299\tiny{±0.0019} & 0.3003\tiny{±0.0026} & 0.1911\tiny{±0.0010} & 0.1700\tiny{±0.0005} & 0.1316\tiny{±0.0050} & 0.1767\tiny{±0.0050} & 0.1123\tiny{±0.0032} & 0.1025\tiny{±0.0028} \\
    & DREAM & 0.2507\tiny{±0.0068} & 0.3255\tiny{±0.0044} & 0.2082\tiny{±0.0044} & 0.1848\tiny{±0.0043} & 0.1469\tiny{±0.0037} & 0.1973\tiny{±0.0037} & 0.1237\tiny{±0.0033} & 0.1118\tiny{±0.0031} \\
    & ABXI & \ul{0.2859}\tiny{±0.0016} & \ul{0.3682}\tiny{±0.0030} & \ul{0.2388}\tiny{±0.0014} & \ul{0.2118}\tiny{±0.0011} & \ul{0.1973}\tiny{±0.0021} & \ul{0.2571}\tiny{±0.0019} & \ul{0.1669}\tiny{±0.0013} & \ul{0.1502}\tiny{±0.0014} \\
    & \textbf{CoRCi} & \textbf{0.2944}\tiny{±0.0020} & \textbf{0.3760}\tiny{±0.0032} & \textbf{0.2472}\tiny{±0.0018} & \textbf{0.2205}\tiny{±0.0014} & \textbf{0.2072}\tiny{±0.0013} & \textbf{0.2697}\tiny{±0.0029} & \textbf{0.1745}\tiny{±0.0016} & \textbf{0.1566}\tiny{±0.0014} \\
    \bottomrule
  \end{tabular}
\vspace{-0.4cm}
\label{tab: rq1}
\end{table*}
\begin{table*}[t]
\caption{Variant performance in MRR (RQ2 and RQ3). Results better than vanilla CoRCi are marked in \colorbox{grey}{Grey}.}
\small
\vspace{-0.3cm}
  \begin{tabular}{cc|ccc|ccc|ccc}
    \toprule
    RQ & Variants & Food & Kitchen & F + K & Beauty & Electronics & B + E & Movie & Book & M + B \\
    \midrule
    - & CoRCi & \textbf{0.1757}\tiny{±0.0023} & \textbf{0.1275}\tiny{±0.0014} & \textbf{0.3032} & \textbf{0.1946}\tiny{±0.0045} & 0.1331\tiny{±0.0013} & \textbf{0.3277} & \textbf{0.2205}\tiny{±0.0014} & 0.1566\tiny{±0.0014} & \textbf{0.3771} \\
    
    \midrule
    \multirow{5}*{RQ2} & V$_{\text{1enc}}$ & 0.1671\tiny{±0.0011} & 0.1206\tiny{±0.0003} & 0.2877 & 0.1876\tiny{±0.0017} & 0.1230\tiny{±0.0028} & 0.3106 & 0.2140\tiny{±0.0022} & 0.1565\tiny{±0.0005} & 0.3705 \\
    
    & V$_{\text{3enc}}$ & 0.1689\tiny{±0.0036} & 0.1213\tiny{±0.0010} & 0.2902 & 0.1891\tiny{±0.0046} & 0.1238\tiny{±0.0018} & 0.3129 & 0.2165\tiny{±0.0023} & 0.1545\tiny{±0.0009} & 0.3710 \\
    
    & V$_{\text{w/o-cat}}$ & 0.1733\tiny{±0.0040} & 0.1249\tiny{±0.0024} & 0.2982 & 0.1921\tiny{±0.0016} & 0.1290\tiny{±0.0020} & 0.3211 & 0.2171\tiny{±0.0011} & 0.1572\tiny{±0.0010} & 0.3743 \\
    
    & V$_{\text{w/o-sg}}$ & 0.1711\tiny{±0.0032} & 0.1199\tiny{±0.0031} & 0.2910 & 0.1900\tiny{±0.0011} & 0.1234\tiny{±0.0009} & 0.3134 & 0.2069\tiny{±0.0016} & 0.1471\tiny{±0.0016} & 0.3540 \\
    
    & V$_{\text{w/o-}\mathcal{L}_{\mathsf{m}}}$ & 0.1569\tiny{±0.0039} & 0.1082\tiny{±0.0034} & 0.2651 & 0.1771\tiny{±0.0047} & 0.1077\tiny{±0.0016} & 0.2848 & 0.1918\tiny{±0.0020} & 0.1356\tiny{±0.0014} & 0.3274 \\
    
    \midrule
    \multirow{8}*{RQ3} & V$_{\text{ff}}$ & 0.1691\tiny{±0.0036} & 0.1224\tiny{±0.0023} & 0.2915 & 0.1890\tiny{±0.0041} & 0.1319\tiny{±0.0012} & 0.3209 & 0.2172\tiny{±0.0018} & 0.1563\tiny{±0.0010} & 0.3735 \\
    
    & V$_{\text{ii}}$ & 0.1696\tiny{±0.0012} & 0.1246\tiny{±0.0017} & 0.2942 & 0.1892\tiny{±0.0055} & 0.1321\tiny{±0.0036} & 0.3213 & 0.2185\tiny{±0.0010} & \cellcolor{grey}\textbf{0.1571}\tiny{±0.0005} & 0.3756 \\
    
    & V$_{\text{fi}}$ & 0.1701\tiny{±0.0027} & 0.1222\tiny{±0.0028} & 0.2923 & 0.1900\tiny{±0.0051} & \cellcolor{grey}0.1333\tiny{±0.0021} & 0.3233 & 0.2154\tiny{±0.0010} & 0.1566\tiny{±0.0010} & 0.3720 \\
    
    & V$_{\text{dt}}$ & 0.1739\tiny{±0.0030} & 0.1234\tiny{±0.0020} & 0.2973 & 0.1894\tiny{±0.0047} & 0.1251\tiny{±0.0022} & 0.3145 & 0.2148\tiny{±0.0022} & 0.1503\tiny{±0.0015} & 0.3651 \\
    
    \cmidrule{2-11}
    
    & M-CoRCi & 0.1695\tiny{±0.0022} & 0.1164\tiny{±0.0039} & 0.2909 & 0.1661\tiny{±0.0059} & \cellcolor{grey}0.1334\tiny{±0.0026} & 0.2995 & 0.1975\tiny{±0.0014} & 0.1444\tiny{±0.0005} & 0.3419 \\
    
    & M-V$_{\text{ff}}$ & 0.1603\tiny{±0.0025} & 0.1119\tiny{±0.0046} & 0.2722 & 0.1593\tiny{±0.0033} & 0.1326\tiny{±0.0012} & 0.2919 & 0.1921\tiny{±0.0017} & 0.1419\tiny{±0.0009} & 0.3340 \\

    & M-V$_{\text{ii}}$ & 0.1636\tiny{±0.0029} & 0.1178\tiny{±0.0025} & 0.2814 & 0.1638\tiny{±0.0040} & 0.1328\tiny{±0.0035} & 0.2966 & 0.1952\tiny{±0.0018} & 0.1446\tiny{±0.0005} & 0.3398 \\

    & M-V$_{\text{fi}}$ & 0.1579\tiny{±0.0016} & 0.1099\tiny{±0.0032} & 0.2678 & 0.1564\tiny{±0.0063} & \cellcolor{grey}\textbf{0.1335}\tiny{±0.0023} & 0.2899 & 0.1901\tiny{±0.0031} & 0.1415\tiny{±0.0011} & 0.3316 \\
    
    & M-V$_{\text{dt}}$ & 0.1676\tiny{±0.0024} & 0.1158\tiny{±0.0040} & 0.2834 & 0.1599\tiny{±0.0050} & 0.1212\tiny{±0.0049} & 0.2911 & 0.1905\tiny{±0.0021} & 0.1348\tiny{±0.0018} & 0.3253 \\
    
    \bottomrule
  \end{tabular}
\label{tab: rq23}
\vspace{-0.3cm}
\end{table*}
%
\subsection{Baselines}
We compare CoRCi to three types of models: single-target CDSR (\textbf{ST-CDSR}) models, dual-target CDSR (\textbf{DT-CDSR}) models, and Sequential Recommendation (\textbf{SR}) models. Furthermore, SR models can be subdivided into three groups based on target settings: single-target (\textbf{ST-SR}), dual-target (\textbf{DT-SR}), and mixed-target (\textbf{MT-SR}).

Specifically, ST-SR models are trained on a single domain at a time. ST-CDSR models are trained on the combined dataset but designate a single domain as the target for each run. Both MT and DT models consider all domains as targets and are trained on a complete cross-domain dataset. The MT-SR models do not distinguish between domains during modeling or loss calculation. In contrast, both DT-SR and DT-CDSR models aggregate per-domain losses. The DT-CDSR model incorporates domain information during training, while DT-SR does not. The baselines are listed as follows: \textbf{SASRec}~\cite{kang2018self} (SR) is a milestone SR model that leverages self-attention to model sequential dynamics. \textbf{BERT4Rec}~\cite{sun2019bert4rec} (SR) leverages the Cloze objective with bidirectional self-attention to model sequential dynamics. \textbf{CD-SASRec}~\cite{alharbi2022cross} (ST-CDSR) aggregates source and target domain sequential dynamics using multiplicative attention and self-attention, respectively, to generate the final representation. \textbf{CD-ASR}~\cite{alharbi2021cross} (ST-CDSR) employs self-attention to extract knowledge from source domains and encode the target-domain sequence; these representations are then aggregated to predict. \textbf{MGCL}~\cite{xu2023multi} (ST-CDSR) designs a multi-view framework with contrastive learning to fuse graphical and sequential information from both mixed-domain and specific-domain perspectives. \textbf{C$^2$DSR}~\cite{cao2022contrastive} (DT-CDSR) establishes contrastive learning by applying random replacement to create negative sequences, reinforcing alignment between mixed- and specific-domain representations. \textbf{DREAM}~\cite{ye2023dream} (DT-CDSR) enhances mixed-domain representations by injecting adaptively extracted knowledge from encoded specific-domain sequential representations. \textbf{ABXI}~\cite{bian2025abxi} (DT-CDSR) unifies the mixed- and specific-domain tasks by employing task-guide alignment and conducts domain-invariant interests adaptation on one shared encoder.

%
\subsection{Implementation Details.}
Following \cite{kang2018self, tang2018personalized, sun2019bert4rec, bian2025abxi}, we adopt leave-one-out strategy. As shown in Figure~\ref{fig: seq}, the last and penultimate interactions of the mixed-domain sequence serve as testing and validation targets, regardless of domain. Metrics are computed per domain. For simplicity and fairness, we follow \cite{kang2018self, tang2018personalized, sun2019bert4rec, bian2025abxi} and sample $N_{\text{mtc}}$ = 999 negative items from the ground truth's domain in metric calculation. Table~\ref{tab: data} reports the number of ground truths per domain in the validation and testing sets. For a fair comparison, unspecified hyperparameter settings follow \cite{bian2025abxi} and are detailed in Table~\ref{tab: hyper}.

\begin{figure}[t]
\centering
\includegraphics[width=\linewidth]{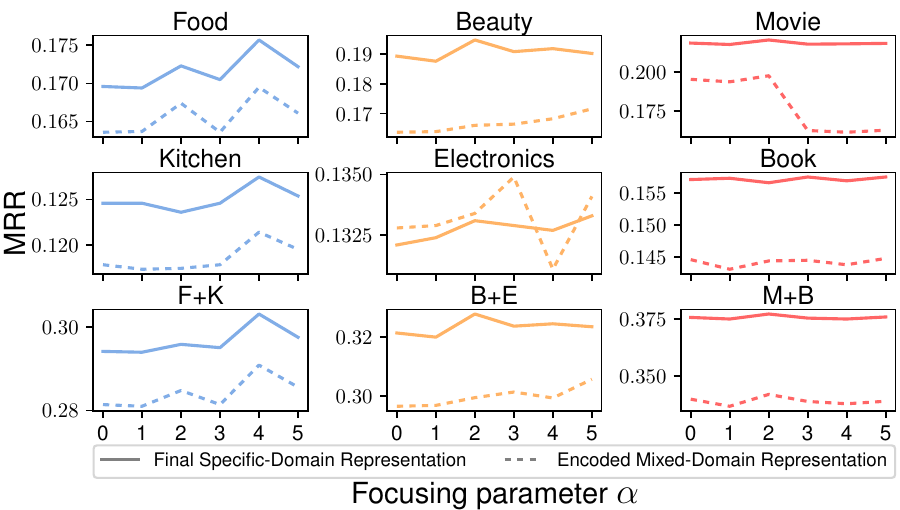}
\setlength{\abovecaptionskip}{-3mm}
\caption{MRR performance of different $\alpha$ (RQ3).}
\vspace{-5mm}
\label{fig: a_focal}
\end{figure}
\begin{figure}[t]
\centering
\includegraphics[width=\linewidth]{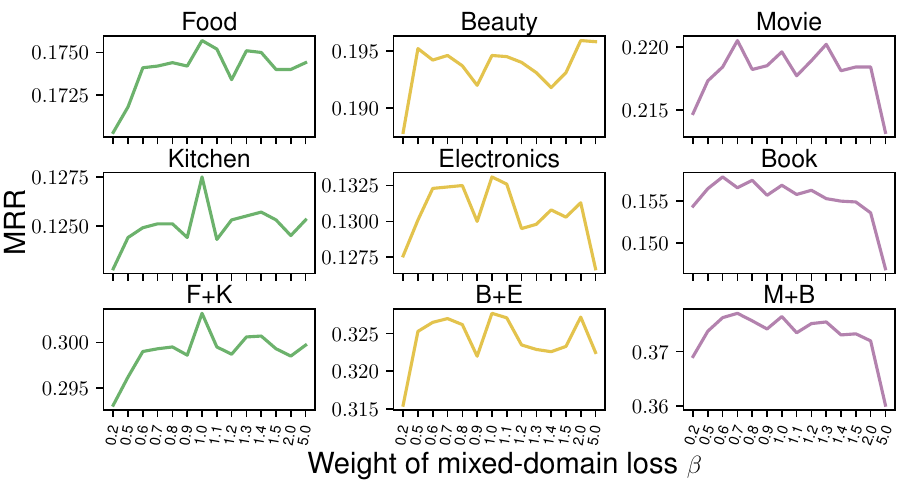}
\setlength{\abovecaptionskip}{-3mm}
\caption{MRR performance of different $\beta$ (RQ4).}
\vspace{-5mm}
\label{fig: w_m}
\end{figure}
%
%
\subsection{Recommendation Performance (\textbf{RQ1})}
For each dataset, we designate the domain with better metrics as the \ul{easy domain} and the other as the \ul{hard domain}. Table~\ref{tab: rq1} summarizes the performance of CoRCi relative to the baselines, from which we derive the following key findings.

Our proposed CoRCi model consistently achieves optimal performance across all evaluated metrics and tasks, with all improvements statistically significant (paired t-tests, $p \leq 0.01$). The average per-domain gains across four metrics are 4.99\% and 6.33\% on AFK, 5.77\% and 11.06\% on ABE, and 3.18\% and 4.68\% on AMB. Notably, CoRCi delivers greater improvements on the hard domains while also making decent progress on easy domains, demonstrating its effectiveness in capturing domain-invariant interests and adaptively reintegrating them into domain-specific modeling.

SASRec-mt and BERT4Rec-mt also perform strongly on the dual-target CDSR task by merely modifying the loss calculation. However, aside from ABXI, no baseline consistently outperforms them across all datasets. Since both CoRCi and ABXI emphasize the capturing of domain-invariant interests, these results validate the importance of this capability in improving CDSR performance.

The DT-CDSR models outperform the ST-CDSR models overall. This is because the dual-target tasks lack noticeable sparsity differences between domains. Therefore, their advantage in transferring knowledge from rich domains to sparse domains is weakened.

\begin{table*}[t]
\small
\setlength{\abovecaptionskip}{2mm}
\caption{MRR Performance on DMBM (RQ5). Paired t-tests show that all improvements are statistically significant ($p$ < 0.01).}
\vspace{-0.1cm}
  \begin{tabular}{c|ccc|ccc|ccc}
    \toprule
    \multirow{2}*{Model} & \multicolumn{3}{c|}{Movie} & \multicolumn{3}{c|}{Book} & \multicolumn{3}{c}{Music} \\
    & HR@10 & NDCG@10 & MRR & HR@10 & NDCG@10 & MRR & HR@10 & NDCG@10 & MRR \\
    \midrule
    ABXI & 0.5921\tiny{±0055} & 0.4534\tiny{±0.0045} & 0.4184\tiny{±0048} & 0.1471\tiny{±0.0061} & 0.0854\tiny{±0.0037} & 0.0763\tiny{±0.0048} & 0.1555\tiny{±0.0173} & 0.0929\tiny{±0.0074} & 0.0845\tiny{±0.0042} \\
    CoRCi & \textbf{0.6216}\tiny{±0.0077} & \textbf{0.4752}\tiny{±0.0047} & \textbf{0.4373}\tiny{±0.0036} & \textbf{0.1746}\tiny{±0.0057} & \textbf{0.1000}\tiny{±0.0068} & \textbf{0.0872}\tiny{±0.0080} & \textbf{0.2049}\tiny{±0.0159} & \textbf{0.1125}\tiny{±0.0060} & \textbf{0.0979}\tiny{±0.0045} \\
    \bottomrule
  \end{tabular}
\label{tab: rq5}
\vspace{-4mm}
\end{table*}
%
\subsection{Ablation Studies (RQ2)}\label{sec: rq2}
In this subsection, we conduct ablation studies to quantitatively evaluate the design choices of CoRCi via five ablation variants:

\noindent\,\textbf{-} V$_{\text{1enc}}$: uses one shared encoder for A and B domains

\noindent\,\textbf{-} V$_{\text{3enc}}$: replace CR$_{\mathsf{M}}$ with SAE$_{\mathsf{M}}$, input $\boldsymbol{E}_{\mathsf{M}}$ as Q, K, and V

\noindent\,\textbf{-} V$_{\text{w/o-cat}}$: remove CAT$_{\mathsf{A}/\mathsf{B}}$

\noindent\,\textbf{-} V$_{\text{w/o-sg}}$: removes stop-gradient operation on $\boldsymbol{H}_{\mathsf{M}}$ 

\noindent\,\textbf{-} V$_{\text{w/o-}\mathcal{L}_\mathsf{M}}$: cancels mixed-domain task on V$_{\text{w/o-sg}}$

We report individual and aggregated domain performance on MRR for each dataset to provide a comprehensive evaluation. As reported in Table~\ref{tab: rq23} (RQ2), the variants V$_{\text{1enc}}$ and V$_{\text{3enc}}$ underperform CoCRi on every metric. V$_{\text{3enc}}$, which deploys three independent encoders, regains only a marginal portion of the lost performance relative to V$_{\text{1enc}}$. These results verify that the proposed SAE–CR hybrid is more effective in modeling domain-invariant interests than either a single shared encoder or fully decoupled encoders. Additionally, the inferior results of V$_{\text{w/o-cat}}$ highlight the crucial role of cross-attention in enabling sequence-level fusion. Likewise, the marked performance declines of V$_{\text{w/o-sg}}$ and V$_{\text{w/o-}\mathcal{L}{\mathsf{M}}}$ verify that both the stop-gradient operation and the standalone mixed-domain objective are indispensable.

%
\subsection{Loss Selections and $\alpha$ in FocalNCE (RQ3)} \label{sec: rq3}
%
We analyze two factors from the perspective of mitigating domain discrepancies: the loss selections and the focusing parameter $\alpha$. 

\noindent\textbf{Loss selections.} We benchmark four ablated variants:

\noindent\,\textbf{-} V$_{\text{ff}}$: FocalNCE to all three losses.

\noindent\,\textbf{-} V$_{\text{ii}}$: InfoNCE to all three losses.
    
\noindent\,\textbf{-} V$_{\text{fi}}$: FocalNCE for $\mathcal{L}_{\mathsf{A}}$ and $\mathcal{L}_{\mathsf{B}}$; InfoNCE for $L_{\mathsf{M}}$.

\noindent\,\textbf{-} V$_{\text{dt}}$: compute $L_{\mathsf{M}}$ separately per domain, same as DT~\cite{cao2022contrastive, ye2023dream, bian2025abxi}.

For each variant V, we report the performance of final specific-domain representations and the performance of encoded mixed-domain representations (M-V, cf. Equation~\ref{eq: cr}). 

In the upper part of Table~\ref{tab: rq23}, we observe that altering FocalNCE and InfoNCE degrades the easy domains of every dataset and the hard domain of AFK. For the hard domains of ABE and AMB, there are statistically insignificant fluctuations.

The lower half of Table~\ref{tab: rq23} reports the performance of the encoded mixed-domain representations. Except for V$_{\text{dt}}$, the results match those of the upper half on AFK, AMB, and the Beauty domain of ABE. On the Electronics domain of ABE, the gap between the two halves is statistically insignificant. This suggests that when CoRCi does not extract useful domain-specific information, it also refrains from injecting harmful noise. These findings suggest that significant mixed-domain performance gains are crucial for enhancing downstream performance.

Notably, V\textsubscript{dt} degenerates consistently across both evaluations. This aligns with the finding in Section~\ref{sec: intro}, which emphasizes that preserving coherent, domain-invariant interests is crucial for improving mixed-domain performance.

\noindent\textbf{FocalNCE focusing parameter $\alpha$.} Figure \ref{fig: a_focal} presents a grid search over $\alpha$. The optimal settings are $\alpha=4$ for AFK, and $\alpha=2$ for both ABE and AMB. In ABE, the Electronics domain is less sensitive to $\alpha$ than the Beauty domain, which matches the findings above. Performance across final specific-domain and encoded mixed-domain representations also follows similar trends. It is worth noting that in the movie domain of AMB, the performance drops sharply when $\alpha\geq3$. This is because user interests in AMB are primarily domain-specific, as discussed in Section~\ref{sec: intro} and Figure~\ref{fig: preliminaries}, while increasing the penalty on same-domain negatives suppresses these interests.
%
\subsection{Impact of Mixed-Domain Weight $\beta$ (RQ4)}
%
Figure \ref{fig: w_m} illustrates the concave relationship between $\beta$ and MRR performance, where too large or too small a value will degrade performance. The optimum occurs at $\beta=1.0$ for AFK and ABE, and $\beta=0.7$ for AMB. The latter corroborates earlier findings in Section~\ref{sec: intro} that AMB is driven more by domain-specific interests. Besides, the Beauty domain exhibits a slight gain at $\beta=5$, but the Electronics domain collapses, confirming that the optimal $\beta$ may differ between paired domains in one dataset. Overall, it is practical to tune the weights between mixed- and specific-domain losses.

%
\subsection{User-Non-Overlapping 3 Domains (RQ5)}\label{sec: rq5}
%
To further assess the generalizability of CoRCi, we extend the experiments to three domains and non-overlapping users, which better reflects real-world applications. Specifically, we utilize the Douban dataset~\cite{zhu2019dtcdr}, incorporating all three available domains (Movie, Book, and Music), namely \textbf{DMBM}. We preprocess the data by removing items with fewer than five occurrences, truncating mixed-domain sequences to the most recent 100 interactions, and filtering users with fewer than five interactions. Users are not required to interact in all domains. Timestamps are recorded at day granularity. For ties within a day, we impose the priority Movie > Book > Music, reflecting the descending interaction volume. This ordering intentionally heightens domain imbalance: Book and Music interactions occur later in the sequence, increasing their likelihood of serving as validation or test ground truths while reducing their effective training lengths.  Table~\ref{tab: data} summarizes the resulting DMBM statistics.

Table~\ref{tab: rq5} contrasts CoRCi with ABXI (the best-performing from RQ1) across three evaluation metrics. CoRCi achieves significant gains over ABXI in all nine domain-metric combinations ($p$ < 0.01).

%
\subsection{Case Study}\label{sec: case}
%
Figure~\ref{fig: cdsr} depicts a shortened snippet from user \textit{A11FFLD0GV82CQ} in the AFK dataset. The complete training sequence is\footnote{Some item pages have been removed, so we identify them based on logged reviews.}: \textit{Granola Cereal} (F) → \textit{Grill Pan} (K) → \textit{Griddle Pan} (K) → \textit{Large Water Dispenser With Standard Filter} (K) → \textit{Fruit and Nut Bar} (F) → \textit{Gourmet Coffee} (F) → \textit{Decaffeinated Green Tea} (F) → \textit{Golden Honey Oat Graham} (F) → \textit{Oat Chocolate Chip Coconut Bar} (F) → \textit{Airtight Container} (K). 

The recent interactions reveal a preference for healthy breakfast snacks (Food) and corresponding storage (Kitchen). The validation and testing items, \textit{Sunrise Crunchy Vanilla Cereal (F)} and \textit{Freshware Silicone Molds (K)}, are thematically consistent. CoRCi ranks them 3rd and 4th, respectively, demonstrating its ability to preserve coherent domain-invariant interests and thus deliver accurate downstream recommendations.

%% file: ch6_conclusion.tex
\vspace{-1mm}
\section{Conclusion}
In this paper, we propose CoRCi, a dual-target CDSR model that employs mixed-domain cross-reconstruction and FocalNCE to mitigate domain discrepancies and maintain domain-invariant interest coherence. Experimental results show that CoRCi outperforms all baselines, including state-of-the-art CDSR counterparts. Future work will investigate a lighter-weight approach that jointly models domain-invariant and domain-specific interests more effectively.